\documentclass{article}

\usepackage[preprint]{neurips_2026}

\usepackage[utf8]{inputenc} 
\usepackage[T1]{fontenc}    
\usepackage{hyperref}       
\usepackage{url}            
\usepackage{booktabs}       
\usepackage{amsfonts}       
\usepackage{nicefrac}       
\usepackage{microtype}      
\usepackage{xcolor}         

\usepackage{amsthm}      
\usepackage{amsmath}     
\usepackage{multirow}    
\usepackage{paralist}    
\usepackage{graphicx}    
\usepackage{algorithm}   
\usepackage{algpseudocode}
\usepackage{tablefootnote} 
\usepackage{subcaption} 
\usepackage{wrapfig} 

\newtheorem*{remark*}{Remark}
\newtheorem{proposition}{Proposition}

\title{Stable Routing for Mixture-of-Experts in Class-Incremental Learning}

\author{%
    Zirui Guo\textsuperscript{1,2} Quan Cheng\textsuperscript{1,2} Da-Wei Zhou\textsuperscript{1,2} Lijun Zhang\textsuperscript{1,2} \\
    \textsuperscript{1}State Key Laboratory of Novel Software Technology, Nanjing University, Nanjing, China \\
    \textsuperscript{2}School of Artificial Intelligence, Nanjing University, Nanjing, China \\
  \texttt{\{guozr,chengq,zhoudw,zhanglj\}@lamda.nju.edu.cn}
}

\begin{document}

\maketitle

\begin{abstract}
  Class-incremental learning (CIL) requires models to learn new classes sequentially while preserving prior knowledge. Recently, approaches that combine pre-trained models with mixture-of-experts (MoE) have received increasing attention in CIL: they typically expand experts during learning and employ a router to assign weights across experts. However, existing MoE methods often overlook \textit{routing drift} induced by expert expansion. Once new experts are introduced, the router may reassign samples from earlier classes to newly added experts, thereby perturbing previously established expert compositions and causing interference even when old experts remain frozen. We argue that expandable MoE in CIL requires two complementary properties: stable old-class routing for knowledge preservation and sufficient capacity utilization for new-class adaptation. To this end, we propose \textbf{\underline{Sta}}ble \textbf{\underline{R}}outing for \textbf{\underline{MoE}} (StaR-MoE), a routing-level framework for expandable MoE in CIL. By incorporating sensitivity-aware routing alignment, StaR-MoE aligns current old-class routing behavior with historical routing distributions through sensitivity-guided constraints. Complementarily, StaR-MoE introduces asymmetric capacity regularization to encourage effective utilization of the expanded expert pool without compromising class-specific routing specialization. Extensive experiments across four standard CIL benchmarks demonstrate that StaR-MoE consistently improves both average and last accuracy over state-of-the-art methods, highlighting the importance of stable routing.
\end{abstract}

\section{Introduction}
Deep neural networks have demonstrated remarkable capabilities when training data are independent and identically distributed~\citep{he2016deep, dosovitskiy2021image}. However, real-world environments are dynamic, requiring the model to learn from a continuous stream of tasks with shifting distributions, known as continual learning (CL)~\citep{wang2024comprehensive,zhou2024class}. The major challenge in CL is catastrophic forgetting, where learning new tasks often interferes with previously acquired knowledge, leading to performance degradation on old tasks~\citep{mccloskey1989catastrophic,mcclelland1995there, french1999catastrophic}.

\begin{wrapfigure}{r}{0.38\linewidth}
    \centering
    \includegraphics[width=\linewidth]{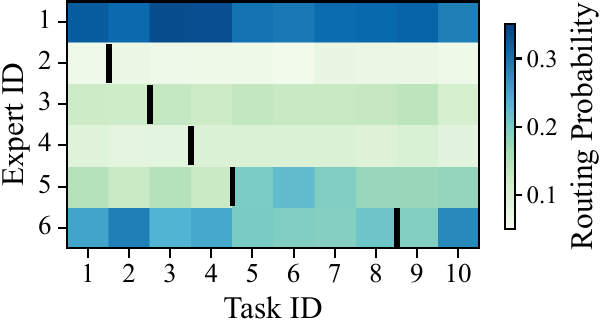}
    \caption{Illustration of \textit{routing drift} in SEMA. Black lines mark expert expansion; non-zero routing probabilities to the left of each marker indicate later experts activated for earlier tasks.}
    \label{fig:router_drift}
    \vspace{-.5em}
\end{wrapfigure}

Among various CL settings, class-incremental learning (CIL) is particularly challenging, as it requires the model to discriminate among all learned classes without access to task identities during inference~\citep{masana2022class,van2022three}.
Recent studies such as SD-LoRA~\citep{wu2025sdlora} and SEMA~\citep{wang2025self} adopt expandable mixture-of-experts (MoE) architectures for CIL, 
where lightweight task-specific modules are progressively introduced and their outputs are aggregated by learnable routing mechanisms. 
However, these methods do not explicitly preserve the routing logic learned for previous classes. As a result, expert expansion can shift old-class routing by normalizing over an expanded expert set, even when historical expert parameters are frozen. As shown in Figure~\ref{fig:router_drift}, newly added experts (e.g., Expert~6) are activated for earlier tasks (e.g., Task~1) despite being introduced later.

We refer to this structural interference problem as \textit{routing drift}, where expert expansion alters the computational pathways of learned classes. 
Routing drift therefore provides a structural explanation for forgetting in expandable MoE: even if historical experts are retained, old-class predictions can still be disrupted when old samples are routed through altered expert compositions, especially when they are assigned to newly added experts that were not optimized for old classes.
To mitigate routing drift in expandable MoE-based CIL, the router should satisfy two complementary properties.
First, the router should maintain the routing behavior of previously learned classes, so that old samples remain unaffected by subsequently added experts.
Second, the router should retain sufficient capacity plasticity for new classes, allowing newly added experts to be effectively utilized without disrupting the established routing paths of old-class samples.

To address routing drift, we propose \textbf{Sta}ble \textbf{R}outing for \textbf{MoE} (StaR-MoE), a routing-level framework that couples historical routing preservation with controlled capacity utilization during expert expansion.
StaR-MoE preserves old-class routing behavior through sensitivity-aware routing alignment (SARA), which aligns current routing distributions with historical class-wise routing targets. Moreover, SARA imposes stronger constraints on router layers that are more sensitive to prediction changes, thereby prioritizing the preservation of routing patterns most critical for retaining old-class computation paths.
We further provide a routing-level theoretical analysis showing that the SARA objective upper-bounds the aggregate routing drift over old classes, thereby offering a principled justification for routing-level preservation.
StaR-MoE further introduces asymmetric capacity regularization (ACR) to maintain capacity plasticity during expert expansion.
ACR regularizes current-task expert loads asymmetrically by penalizing only overloaded experts, thus encouraging the router to exploit the expanded expert pool for new classes without compromising routing specialization.


In summary, our main contributions are as follows:
\begin{compactitem}
    \item We identify \textit{routing drift} as a structural interference problem in expandable MoE-based CIL, where newly added experts can alter the computational pathways of previously learned classes even when historical experts are frozen.

    \item We propose StaR-MoE, a routing-level framework for mitigating routing drift by coupling historical routing preservation with controlled capacity utilization. It introduces sensitivity-aware routing alignment for preserving old-class routing and asymmetric capacity regularization to encourage expert utilization.

    \item We conduct extensive experiments on multiple CIL benchmarks, demonstrating the effectiveness of StaR-MoE in maintaining stable routing and accommodating new tasks.
\end{compactitem}

\section{Related Work}

\subsection{Continual Learning} 
Continual Learning (CL) aims to learn a sequence of tasks from non-stationary data streams without catastrophic forgetting~\citep{mccloskey1989catastrophic,masana2022class,wang2024comprehensive,zhou2024class}. Existing CL methods can be broadly classified into three primary categories: regularization-based, replay-based, and architecture-based approaches. Regularization-based methods~\citep{kirkpatrick2017overcoming, li2017learning, zenke2017continual, aljundi2018memory} incorporate penalty terms into the loss function to constrain the updates of important parameters for previous tasks. Replay-based methods~\citep{rebuffi2017icarl, lopez2017gradient, buzzega2020dark, choi2021dual, sun2022exploring} mitigate forgetting by retaining a buffer of raw data or training a generative model. More recently, several approaches~\citep{zhang2023slca, wang2023hierarchical, li2025addressing} alleviate memory overhead by modeling feature representations with class-wise Gaussian distributions. Architecture-based methods~\citep{yoon2018lifelong, li2019learn, sokar2021spacenet, liang2023adaptive} dynamically expand the model capacity to accommodate new tasks. By isolating task-specific parameters, these approaches can retain acquired knowledge.

\subsection{Continual Learning with Pre-trained Models}
While traditional approaches primarily focus on models trained from scratch, recent strategies that pair a frozen pre-trained backbone with trainable lightweight components have demonstrated effectiveness in CL~\citep{wang2022learning, smith2023coda, huang2024class}. Specifically, methods such as L2P~\citep{wang2022learning} and DualPrompt~\citep{wang2022dualprompt} integrate pre-trained models with prompt-tuning strategies~\citep{jia2022visual}, learning to retrieve relevant prompts to adapt the frozen representations for different tasks. However, the absence of task identities in CIL makes such retrieval prone to errors. Mismatched prompts may activate inappropriate task-specific adaptation paths and degrade old-class predictions~\citep{sun2025mos}. CODA-Prompt~\citep{smith2023coda} alleviates rigid prompt selection by dynamically combining prompt components, further enhancing adaptation flexibility. Moving beyond prompt-tuning, recent works such as InfLoRA~\citep{liang2024inflora} incorporate specialized Low-Rank Adaptation (LoRA)~\citep{hu2022lora} components and expand them during training. Similarly, approaches such as EASE~\citep{zhou2024expandable} and APER~\citep{zhou2025revisiting} employ adapter modules~\citep{chen2022adaptformer} for parameter-efficient fine-tuning.

\subsection{Mixture of Experts in Continual Learning}
To further scale model capacity while reducing the reliance on hard task-specific retrieval, recent research has adopted the MoE architecture, where multiple experts are softly composed through routing. Typically, an MoE comprises multiple experts and a router that assigns routing weights to aggregate their outputs~\citep{shazeer2017outrageously, fedus2022switch, huai2025cl, le2026one}, making the router a critical component for controlling expert utilization. For instance, MoE-Adapters~\citep{yu2024boosting} introduces a dedicated router for each task over a fixed set of experts, and selects the most relevant router during inference. SD-LoRA~\citep{wu2025sdlora} utilizes task-specific LoRA components with decoupled direction and magnitude, and employs learnable scalar weights to combine these components. To address linear expert growth, SEMA~\citep{wang2025self} detects distribution shifts to adaptively decide when to expand. However, existing MoE-based CL methods mainly focus on expert construction or expansion, which leaves routing drift induced by continual expert expansion largely unaddressed. In contrast, StaR-MoE explicitly mitigates routing drift by preserving historical routing behavior, retaining stable routing for old classes while enabling the model to effectively learn new classes.

\section{Preliminaries}
We first introduce the CIL setting and the expandable MoE architecture used throughout the paper.
\subsection{Class-Incremental Learning}
In the CIL setting, the model learns from a continuous data stream where task identities are unavailable at inference time~\citep{wang2022dualprompt, wu2025sdlora}. Formally, the training process consists of $T$ sequential tasks $\{\mathcal{T}_1, \dots, \mathcal{T}_T\}$. Each task $\mathcal{T}_t$ provides a dataset $\mathcal{D}_t = \{(x_n, y_n)\}_{n=1}^{N_t}$ with $N_t$ instances, where $x_n$ is the input image and $y_n \in \mathcal{C}_t$ is the corresponding label. The label sets are mutually disjoint, i.e., $\mathcal{C}_i \cap \mathcal{C}_j = \emptyset$ for $i \neq j$. At step $t$, only data from $\mathcal{D}_t$ are accessible for training, and the model $f_{\theta_t}$ parameterized by $\theta_t$ is optimized on the current dataset via the classification loss:
\begin{equation}
\label{eq:cur_loss}
    \mathcal{L}_{\mathrm{cur}}(\theta_t;\mathcal{D}_t) = \mathbb{E}_{(x, y) \sim \mathcal{D}_t} [\ell(f_{\theta_t}(x), y)],
\end{equation}
where $\mathbb{E}[\cdot]$ denotes the expectation, and $\ell(\cdot, \cdot)$ is a per-sample loss function such as cross-entropy. The overall objective is to minimize the expected risk over all classes observed up to step $t$, i.e., $\mathcal{C}_{1:t} = \bigcup_{i=1}^t \mathcal{C}_i$, which requires the model to learn new tasks while mitigating forgetting on previous tasks whose data are unavailable.

\subsection{Parallel Adapter as Expert}
A Vision Transformer (ViT)~\citep{dosovitskiy2021image} consists of multiple transformer layers, each containing a multi-head self-attention module and a multi-layer perceptron (MLP) block. Following prior work~\citep{chen2022adaptformer, he2022towards}, we freeze the ViT backbone and insert lightweight adapters as experts in parallel with the MLP blocks. Let $\mathbf{x}^l \in \mathbb{R}^{d}$ denote the input feature to the MLP block at the $l$-th layer. The adapter function $A^l (\mathbf{x}^l)$ comprises a down-projection $\mathbf{W}_{\mathrm{down}}^l \in \mathbb{R}^{d \times r}$ that maps features to a bottleneck dimension $r$, followed by a ReLU activation~\citep{nair2010rectified} and an up-projection $\mathbf{W}_{\mathrm{up}}^l \in \mathbb{R}^{r \times d}$. For brevity, we omit the layer index $l$ in the following formulations. The adapter structure is then defined as:
\begin{equation}
    A(\mathbf{x}) = \mathrm{ReLU}(\mathbf{x} \mathbf{W}_{\mathrm{down}})\mathbf{W}_{\mathrm{up}}.
\end{equation}

\subsection{Mixture-of-Experts with Top-$k$ Routing}
In each expandable MoE layer, a new adapter-based expert is added for each new task, and the router is expanded accordingly.
Let $E_{\mathrm{total}}$ denote the number of available experts at the current task.
Given the \texttt{[CLS]} token $\boldsymbol{z}\in\mathbb{R}^{d}$ as the router input, the routing logits are computed as:
\begin{equation}
    \boldsymbol{h} = \boldsymbol{z}\mathbf{W}_{\mathrm{router}},
\end{equation}
where $\mathbf{W}_\mathrm{router} \in\mathbb{R}^{d\times E_\mathrm{total}}$ is the learnable routing matrix. Following the standard Top-$k$ MoE routing~\citep{shazeer2017outrageously}, we define
$\mathcal{K}=\mathrm{Top}\text{-}k(\boldsymbol{h})$ as the selected expert set, with
$|\mathcal{K}|=\min(k,E_{\mathrm{total}})$.
The sparse routing probability is defined as:
\begin{equation}
    G(\boldsymbol{z})_j =
    \begin{cases}
        \dfrac{\exp(h_j)}{\sum_{m\in\mathcal{K}}\exp(h_m)}, & j\in\mathcal{K}, \\[1.5ex]
        0, & j\notin\mathcal{K}.
    \end{cases}
    \label{eq:sparse_gate}
\end{equation}
The MoE output is computed as:
\begin{equation}
    \mathbf{x}_{\mathrm{out}}
    =
    \mathrm{MLP}(\mathbf{x})
    +
    \sum_{j\in\mathcal{K}} G(\boldsymbol{z})_j A_j(\mathbf{x}),
    \label{eq:moe_output}
\end{equation}
where $A_j(\cdot)$ denotes the $j$-th adapter expert. In CIL, the expert pool and routing matrix expand over tasks.
Without explicit constraints, expert expansion may change the routing distribution of earlier classes and alter their computational pathways.

\begin{figure*}[t]
  \centering
  \includegraphics[width=\textwidth]{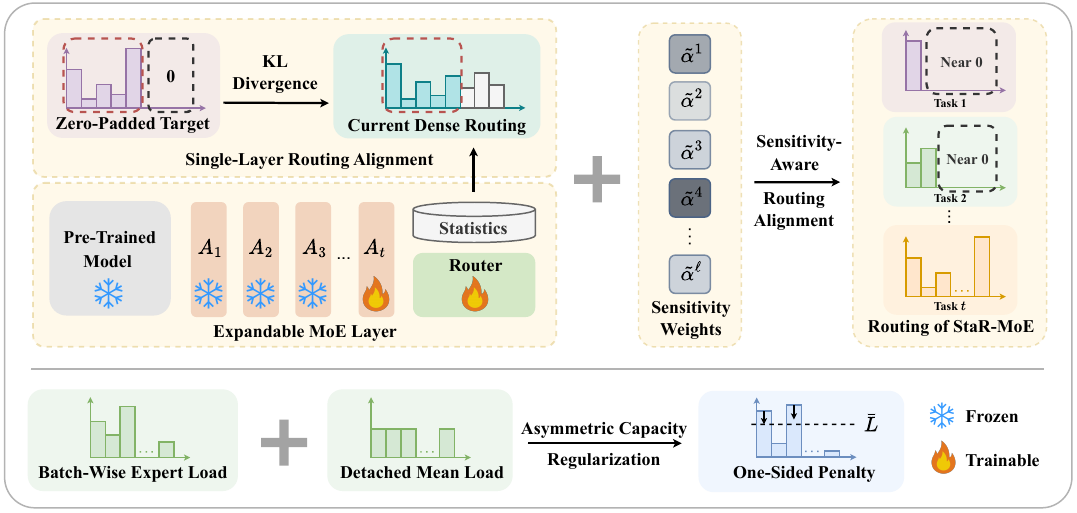} 
  \caption{Overview of the StaR-MoE framework.}
  \label{fig:framework}
\end{figure*}

\section{Methodology}
\label{sec:method}
In this section, we present StaR-MoE in detail. As illustrated in Figure~\ref{fig:framework}, StaR-MoE consists of two complementary components: sensitivity-aware routing alignment (SARA), which preserves historical routing for old classes, and asymmetric capacity regularization (ACR), which encourages effective capacity utilization under dynamic expert expansion. We first introduce SARA and then describe ACR.

\subsection{Sensitivity-Aware Routing Alignment}
\label{sec:sara}
To stabilize the routing for learned classes, we propose sensitivity-aware routing alignment (SARA), which maintains the historical routing behavior of old classes while focusing the alignment constraint on routers that are more sensitive to the current task optimization.

\textbf{Single-layer routing alignment.}
We first describe the routing alignment objective for a single MoE layer. Given the router input $\boldsymbol{z}$ and routing matrix $\mathbf{W}_{\mathrm{router}}$, the dense routing distribution vector is defined as:
\begin{equation}
    \boldsymbol{P}(\boldsymbol{z}) = \mathrm{Softmax}(\boldsymbol{z}\mathbf{W}_{\mathrm{router}}).
\end{equation}
We use the dense routing distribution for alignment instead of the sparse Top-$k$ routing used in the forward pass. Dense routing provides gradients to all expert logits and directly suppresses newly added experts for old classes. As Top-$k$ selection is induced by the same logits, stabilizing dense distributions also stabilizes sparse expert selection.

Once class $c$ has been learned, we derive its class-wise routing anchor to summarize its historical expert preference. Specifically, given the router inputs $\{\boldsymbol{z}_{c,i}\}_{i=1}^{N_c}$ of class $c$, we compute the average routing logits as $\bar{\boldsymbol{l}}_{c} = \frac{1}{N_c} \sum_{i=1}^{N_c} \boldsymbol{z}_{c,i}\mathbf{W}_{\mathrm{router}}$, where $N_c$ is the number of training samples in class $c$. The corresponding historical routing distribution is $\bar{\boldsymbol{P}}_{c} = \mathrm{Softmax}(\bar{\boldsymbol{l}}_{c})$. When new experts are added in later tasks, we construct a zero-padded target distribution for class $c$:
\begin{equation}
    \hat{\boldsymbol{P}}_{c}
    =
    \mathrm{Concat}
    \left(
    \bar{\boldsymbol{P}}_{c},
    \mathbf{0}_{\mathrm{new}}
    \right),
    \label{eq:padding}
\end{equation}
where $\mathbf{0}_{\mathrm{new}}$ denotes zero probabilities assigned to newly added experts. This target distribution explicitly encodes the temporal constraint of expert expansion: experts introduced after class $c$ was learned should not be activated for class $c$.

Since old training data are unavailable during the learning process, SARA approximates router inputs of old classes effectively through class-wise statistics. While prior works~\citep{zhu2021prototype, zhang2023slca} typically use class-wise Gaussian statistics at the final layer for classifier retraining, SARA instead models router-input distributions across MoE layers, with empirical visualizations provided in Appendix~\ref{app:router_input_visualization}. For each old class $c$, we model the router-input distribution as $\mathcal{N}(\boldsymbol{\mu}_{c}, \boldsymbol{\Sigma}_{c})$, where $\boldsymbol{\mu}_{c}$ and $\boldsymbol{\Sigma}_{c}$ are the mean and diagonal covariance of the router inputs from class $c$. During later tasks, we sample synthetic router inputs $\tilde{\boldsymbol{z}}_{c} \sim \mathcal{N}(\boldsymbol{\mu}_{c}, \boldsymbol{\Sigma}_{c})$ and feed them only into the router to obtain the current dense routing distribution $\boldsymbol{P}(\tilde{\boldsymbol{z}}_{c})$. The single-layer routing alignment loss is then defined using the Kullback-Leibler (KL) divergence as:
\begin{equation}
\label{eq:align_loss}
\begin{split}
    \mathcal{L}_{\mathrm{align}}
    &=
    \sum_{c \in \mathcal{C}_{\mathrm{old}}}
    \mathbb{E}_{\tilde{\boldsymbol{z}}_{c} \sim \mathcal{N}(\boldsymbol{\mu}_{c}, \boldsymbol{\Sigma}_{c})}
    \left[
    D_{\mathrm{KL}}
    \left(
    \hat{\boldsymbol{P}}_{c}
    \parallel
    \boldsymbol{P}(\tilde{\boldsymbol{z}}_{c})
    \right)
    \right] \\
    &=
    \sum_{c \in \mathcal{C}_{\mathrm{old}}}
    \mathbb{E}_{\tilde{\boldsymbol{z}}_{c}}
    \left[
    \sum_{j=1}^{E_{\mathrm{total}}}
    \hat{\boldsymbol{P}}_{c,j}
    \log
    \left(
    \frac{
    \hat{\boldsymbol{P}}_{c,j}
    }{
    \boldsymbol{P}(\tilde{\boldsymbol{z}}_{c})_j
    }
    \right)
    \right],
\end{split}
\end{equation}
where $\mathcal{C}_{\mathrm{old}}$ denotes the set of previously learned classes, and $E_{\mathrm{total}}$ is the total number of experts. By matching $\hat{\boldsymbol{P}}_{c}$ and $\boldsymbol{P}(\tilde{\boldsymbol{z}}_{c})$, the alignment loss preserves the relative expert preference learned for each old class and suppresses the routing probability assigned to newly introduced experts.

\textbf{Sensitivity-aware weighting.}
While Eq.~\eqref{eq:align_loss} stabilizes the router in a single MoE layer, StaR-MoE introduces routers across multiple transformer layers. Applying the same alignment strength to all routers can be suboptimal, as routing drift in different layers may affect the final prediction to different extents. Over-constraining less sensitive routers may impair new knowledge acquisition, while under-constraining highly sensitive routers may induce severe forgetting of previously acquired knowledge. Therefore, SARA adaptively weights the alignment objective according to router sensitivity.

For the $\ell$-th MoE layer, we denote its routing matrix by $\mathbf{W}_{\mathrm{router}}^{\ell}$ and define its alignment loss $\mathcal{L}_{\mathrm{align}}^{\ell}$ following Eq.~\eqref{eq:align_loss}. Before training on the current task, SARA estimates the sensitivity of each router with a lightweight one-batch gradient proxy. Specifically, SARA uses only a single mini-batch $\mathcal{B}_{t}$ from the current task, without storing raw samples from previous tasks. SARA then performs one forward and one backward pass using the classification loss $\mathcal{L}_{\mathrm{cur}}$, without updating any parameters. The sensitivity score of the $\ell$-th router is defined as the gradient norm with respect to its routing matrix:
\begin{equation}
    s^{\ell}
    =
    \left\|
    \nabla_{\mathbf{W}_{\mathrm{router}}^{\ell}}
    \mathcal{L}_{\mathrm{cur}}(\theta_t;\mathcal{B}_{t})
    \right\|_F .
\end{equation}
The sensitivity scores are normalized across all MoE layers as:
\begin{equation}
    \alpha^{\ell}
    =
    \frac{\exp(s^{\ell})}
    {
    \sum_{\ell'\in\mathcal{L}_{\mathrm{moe}}}
    \exp(s^{\ell'})
    },
\end{equation}
where $\mathcal{L}_{\mathrm{moe}}$ denotes the set of MoE layers. To avoid an overly concentrated constraint on a single layer, we interpolate the sensitivity weights with uniform weights:
\begin{equation}
    \tilde{\alpha}^{\ell}
    =
    \gamma \alpha^{\ell}
    +
    (1-\gamma)
    \frac{1}{|\mathcal{L}_{\mathrm{moe}}|},
\end{equation}
where $\gamma\in[0,1]$ controls the strength of sensitivity-aware weighting. The final SARA objective is:
\begin{equation}
\label{eq:sara_loss}
    \mathcal{L}_{\mathrm{SARA}}
    =
    \sum_{\ell\in\mathcal{L}_{\mathrm{moe}}}
    \tilde{\alpha}^{\ell}
    \mathcal{L}_{\mathrm{align}}^{\ell}.
\end{equation}

\textbf{Routing drift bound.}
We further relate the alignment loss to routing drift. Consider the set of old classes $\mathcal{C}_{\mathrm{old}}$ with cardinality $N = |\mathcal{C}_{\mathrm{old}}|$. For the $\ell$-th MoE layer and each class $c \in \mathcal{C}_{\mathrm{old}}$, let $\hat{\boldsymbol{P}}_{c}^{\ell}$ denote the zero-padded target distribution, and let $\boldsymbol{P}^{\ell}(\tilde{\boldsymbol{z}}_{c}^{\ell})$ denote the current dense routing distribution on the synthesized router input $\tilde{\boldsymbol{z}}_{c}^{\ell}$. We define the layer-wise routing drift as:
\begin{equation}
    \Delta^{\ell}
    =
    \sum_{c \in \mathcal{C}_{\mathrm{old}}}
    \mathbb{E}_{\tilde{\boldsymbol{z}}_{c}^{\ell}}
    \left[
    \left\|
    \hat{\boldsymbol{P}}_{c}^{\ell}
    -
    \boldsymbol{P}^{\ell}(\tilde{\boldsymbol{z}}_{c}^{\ell})
    \right\|_1
    \right].
\end{equation}
Since SARA assigns different alignment strengths to different layers, we define the sensitivity-weighted routing drift across all MoE layers as:
\begin{equation}
    \Delta_{\mathrm{total}}
    =
    \sum_{\ell\in\mathcal{L}_{\mathrm{moe}}}
    \tilde{\alpha}^{\ell}
    \Delta^{\ell}.
\end{equation}

\begin{proposition}[Bound on Routing Drift]
\label{prop:sara_drift_bound}
If the SARA objective satisfies $\mathcal{L}_{\mathrm{SARA}} \leq \xi$, then the sensitivity-weighted routing drift is bounded by:
\begin{equation}
    \Delta_{\mathrm{total}}
    \leq
    \sqrt{2N\xi}.
\end{equation}
\end{proposition}

\begin{remark*}
Proposition~\ref{prop:sara_drift_bound} establishes a worst-case bound on the routing drift. When the divergence is concentrated on a small subset of classes or layers, the actual routing drift is typically much smaller than the bound. The proof is provided in Appendix~\ref{app:proof_propo_1}.
\end{remark*}

\subsection{Asymmetric Capacity Regularization}
\label{sec:acr}

While SARA stabilizes routing for old classes, relying on routing alignment alone may make the router overly biased toward historical experts. Since the zero-padded targets in SARA assign zero probability to newly added experts for old classes, the router may become biased toward historical experts and fail to fully utilize the expanded capacity for the current task. Thus, we introduce asymmetric capacity regularization (ACR), which encourages effective capacity utilization under dynamic expert expansion.

Let $\mathcal{B}$ denote a batch of router inputs derived from the current task, and let $\mathcal{K}$ be the Top-$k$ expert set selected by the router for input $\boldsymbol{z}$. For each expert $j\in\{1,\dots,E_{\mathrm{total}}\}$, we compute its batch-level load as:
\begin{equation}
    L_j
    =
    \sum_{\boldsymbol{z}\in\mathcal{B}}
    \mathcal{S}
    \left(
    j\in \mathcal{K}
    \mid
    \boldsymbol{z}
    \right),
\end{equation}
where $\mathcal{S}(j\in \mathcal{K} \mid \boldsymbol{z})$ denotes the smooth probability that expert $j$ is selected into the Top-$k$ set for router input $\boldsymbol{z}$. The detailed formulation of $\mathcal{S}(\cdot)$ is provided in Appendix~\ref{app:smooth_load}. The load $L_j$ measures the expected selection frequency of expert $j$ on current router inputs.

Conventional load-balancing regularization in MoE~\citep{shazeer2017outrageously, fedus2022switch} encourages all experts to be used uniformly. However, such a symmetric objective is designed for static expert sets and does not account for dynamic expert expansion in CIL. In expandable MoE, previously learned experts should store old knowledge, while newly added experts should provide capacity for new-task adaptation. Forcing old and new experts toward uniform usage may therefore conflict with established expert specialization and amplify routing drift.

To address this issue, ACR complements SARA with an asymmetric capacity penalty. We define the detached mean load as:
\begin{equation}
    \bar{L}
    =
    \mathrm{sg}
    \left(
    \frac{1}{E_{\mathrm{total}}}
    \sum_{j=1}^{E_{\mathrm{total}}}
    L_j
    \right),
\end{equation}
where $\mathrm{sg}(\cdot)$ denotes the stop-gradient operation. Instead of penalizing all experts uniformly, ACR only penalizes experts whose loads exceed the mean:
\begin{equation}
\label{eq:acr_loss}
    \mathcal{L}_{\mathrm{ACR}}
    =
    \frac{
    \frac{1}{E_{\mathrm{total}}}
    \sum_{j=1}^{E_{\mathrm{total}}}
    \left[
    \max(L_j-\bar{L},0)
    \right]^2
    }
    {
    \bar{L}^2+\epsilon
    }.
\end{equation}
Here, $\epsilon$ is a small constant for numerical stability. This one-sided formulation prevents a small subset of experts from dominating the routing allocation on the current task, while allowing non-uniform expert usage to preserve expert specialization. ACR improves the utilization of newly expanded capacity by discouraging excessive load concentration, while preserving routing specialization.

\textbf{Total Objective.}
The final training objective of StaR-MoE is given by:
\begin{equation}
    \mathcal{L}_{\mathrm{total}}
    =
    \mathcal{L}_{\mathrm{cur}}
    +
    \lambda_{\mathrm{SARA}}
    \mathcal{L}_{\mathrm{SARA}}
    +
    \lambda_{\mathrm{ACR}}
    \mathcal{L}_{\mathrm{ACR}},
\end{equation}
where $\lambda_{\mathrm{SARA}}$ and $\lambda_{\mathrm{ACR}}$ control the strength of routing alignment and capacity regularization, respectively. The complete StaR-MoE algorithm is presented in Appendix~\ref{app:alg}.

\section{Experiments}
\label{sec:exp}
In this section, we evaluate StaR-MoE on multiple class-incremental learning benchmarks. We first describe the experimental settings, then present benchmark comparisons, ablation studies, and further analysis on routing drift.

\subsection{Experimental Settings}

\textbf{Datasets.} Following previous work~\citep{smith2023coda,liang2024inflora,wang2025self}, we evaluate StaR-MoE on four standard class-incremental learning benchmarks: ImageNet-R~\citep{hendrycks2021many}, ImageNet-A~\citep{hendrycks2021natural}, CIFAR-100~\citep{krizhevsky2009learning}, and VTAB~\citep{zhai2019large}. We evaluate ImageNet-R under three different settings by splitting it into 5, 10, and 20 tasks, with each task containing 40, 20, and 10 classes, respectively. For ImageNet-A, we divide it into 10 tasks, each containing 20 classes. Similarly, CIFAR-100 is split into 10 tasks, each containing 10 classes. Finally, for VTAB, which comprises 50 classes across five domains, we organize it into 5 tasks with 10 classes each.

\textbf{Evaluation metrics.}  Two standard continual learning metrics widely used in recent studies are employed for evaluation~\citep{wang2022learning,wu2025sdlora}. The first metric, last accuracy $\mathcal{A}_T$, measures the average classification accuracy across all tasks after the complete training sequence. The second metric, average accuracy $\bar{\mathcal{A}}$, evaluates the overall performance throughout the incremental process. It is defined as $\bar{\mathcal{A}} = \frac{1}{T} \sum_{t=1}^{T} \mathcal{A}_t$, where $\mathcal{A}_t$ is the accuracy on all learned classes after task $t$. These two metrics capture both the model's ability to learn new tasks and to retain knowledge of previously learned ones.

\textbf{Baselines and implementation details.} We compare StaR-MoE against state-of-the-art pre-trained ViT-based methods. Prompt-based methods include L2P~\citep{wang2022learning}, DualPrompt~\citep{wang2022dualprompt}, and CODA-Prompt~\citep{smith2023coda}. Statistics-based methods include SimpleCIL~\citep{zhou2025revisiting} and APER~\citep{zhou2025revisiting}. Dynamically expandable methods include InfLoRA~\citep{liang2024inflora}, SD-LoRA~\citep{wu2025sdlora}, and SEMA~\citep{wang2025self}. Comparing against these diverse approaches allows us to comprehensively evaluate the effectiveness of StaR-MoE.
In addition to the ViT-B/16~\citep{dosovitskiy2021image} pre-trained on ImageNet-21K and fine-tuned on ImageNet-1K using supervised learning, we also evaluate a self-supervised ViT-B/16 obtained with DINO~\citep{caron2021emerging}. Details of the experiments are provided in Appendix~\ref{app:detail}.

\subsection{Benchmark Comparison}
\label{sec:benchmark_comparison}
Table~\ref{tab:main_results} reports the comparison with state-of-the-art ViT-based CIL methods. StaR-MoE consistently achieves the best performance across all benchmarks and task splits. On ImageNet-R, StaR-MoE obtains $\bar{\mathcal{A}}=86.09\%$, $84.96\%$, and $82.90\%$ under the 5-task, 10-task, and 20-task settings, respectively, showing robust performance as the task sequence becomes longer. 
The improvements are more pronounced on ImageNet-A and VTAB, which involve stronger distribution shifts. On ImageNet-A, StaR-MoE achieves $\bar{\mathcal{A}}=71.60\%$ and $\mathcal{A}_T=61.75\%$, improving over SEMA by $4.57$ and $3.69$ points, respectively. On VTAB, StaR-MoE obtains $\bar{\mathcal{A}}=94.43\%$ and $\mathcal{A}_T=93.79\%$, surpassing the best baseline in each metric by $3.50$ and $5.25$ points. These results indicate that stable routing is particularly beneficial under distribution-shifted incremental scenarios.

\begin{figure}[t]
    \centering
    \begin{subfigure}{0.32\textwidth}
        \centering
        \includegraphics[width=0.98\linewidth]{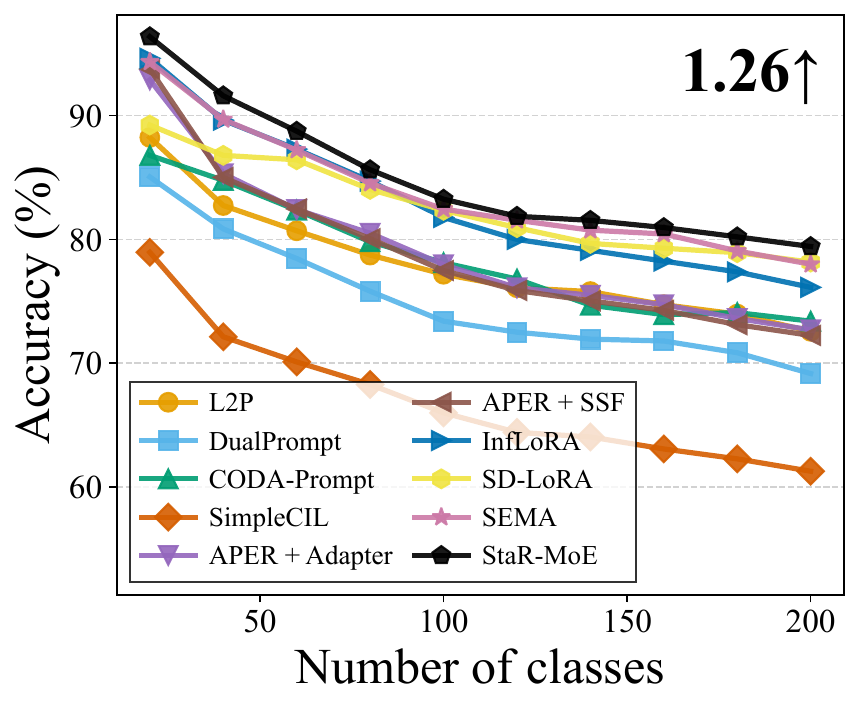} 
        \caption{ImageNet-R (10 tasks)} 
        \label{fig:inr_10}
    \end{subfigure}
    \hfill
    \begin{subfigure}{0.32\textwidth}
        \centering
        \includegraphics[width=0.98\linewidth]{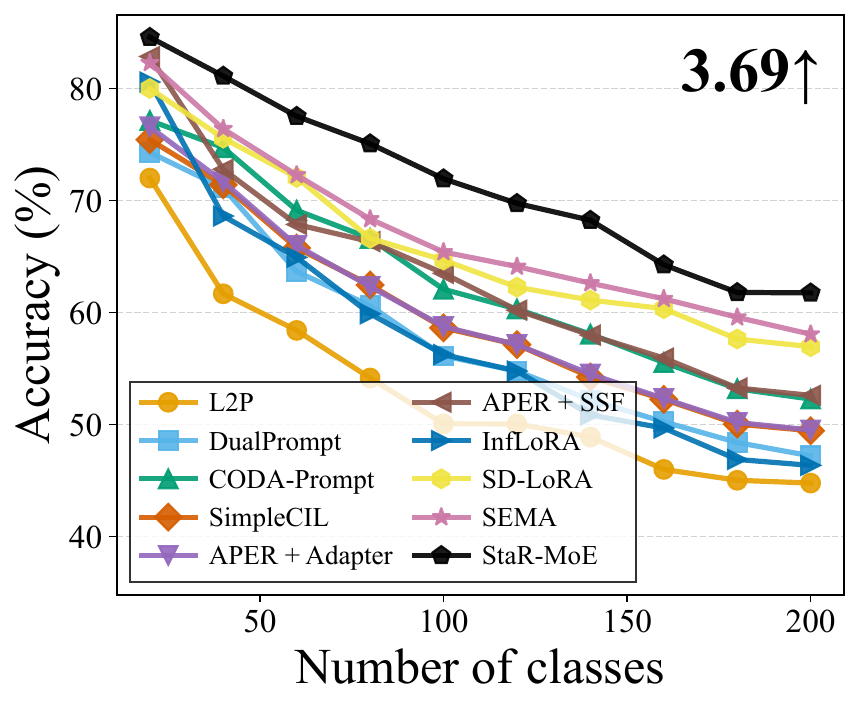} 
        \caption{ImageNet-A} 
        \label{fig:ina}
    \end{subfigure}
    \hfill
    \begin{subfigure}{0.32\textwidth}
        \centering
        \includegraphics[width=0.98\linewidth]{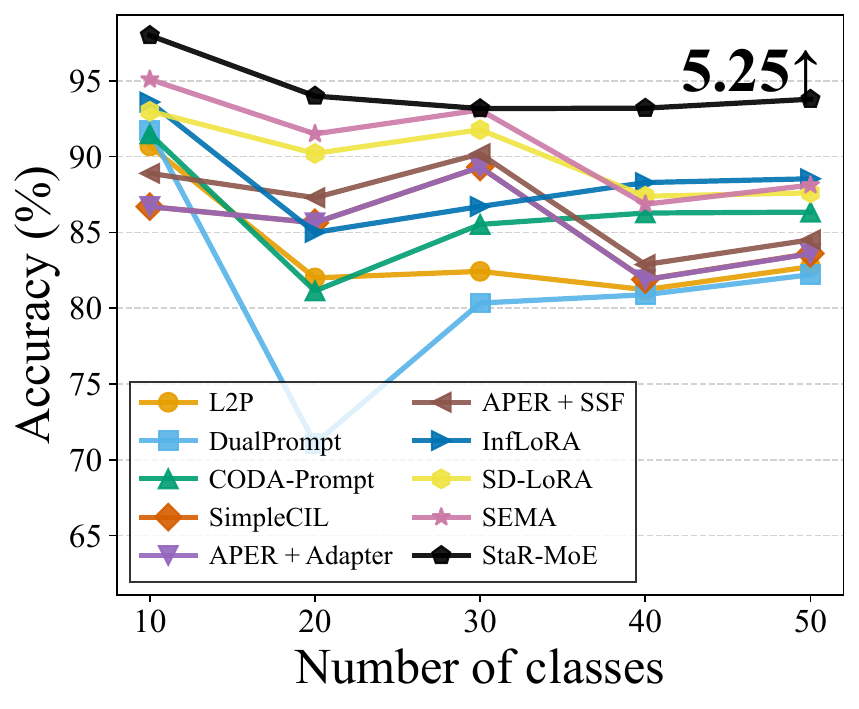} 
        \caption{VTAB}
        \label{fig:vtab}
    \end{subfigure}
    
    \caption{Performance curves of different methods on ImageNet-R, ImageNet-A, and VTAB. We annotate StaR-MoE's final-stage accuracy gain over the runner-up method in the top-right corner of each subfigure.}
    \label{fig:accuracy_curves}
\end{figure}

Figure~\ref{fig:accuracy_curves} further shows the task-wise accuracy curves on representative benchmarks. StaR-MoE maintains a consistent advantage throughout the incremental process, and the final-stage gains annotated in each subfigure confirm its superiority over the runner-up methods. This suggests that mitigating routing drift helps preserve previously learned knowledge while adapting to new classes.

\begin{table}[t]
\caption{Comparison with state-of-the-art ViT-based methods. All models adopt ViT-B/16-IN1K as the backbone. IN-R stands for ImageNet-R. $\bar{\mathcal{A}}$ denotes average accuracy, and $\mathcal{A}_T$ denotes last accuracy. The best results are highlighted in \textbf{bold}.}
\label{tab:main_results}
\centering
\resizebox{\textwidth}{!}{%
\begin{tabular}{lcccccccccccc}
\toprule
\multirow{2}{*}{Method} & \multicolumn{2}{c}{CIFAR-100} & \multicolumn{2}{c}{5-Task IN-R} & \multicolumn{2}{c}{10-Task IN-R} & \multicolumn{2}{c}{20-Task IN-R} & \multicolumn{2}{c}{ImageNet-A} & \multicolumn{2}{c}{VTAB}\\
 & $\bar{\mathcal{A}}$ & $\mathcal{A}_T$ & $\bar{\mathcal{A}}$ & $\mathcal{A}_T$ & $\bar{\mathcal{A}}$ & $\mathcal{A}_T$ & $\bar{\mathcal{A}}$ & $\mathcal{A}_T$ & $\bar{\mathcal{A}}$ & $\mathcal{A}_T$ & $\bar{\mathcal{A}}$ & $\mathcal{A}_T$\\
\midrule
L2P & 87.18 & 81.85 & 77.40 & 73.59 & 66.97 & 62.72 & 70.67 & 62.90 & 51.40 & 42.94 & 81.19 & 80.83 \\
DualPrompt & 87.36 & 82.30 & 76.39 & 72.29 & 72.83 & 66.75 & 62.33 & 61.97 & 54.68 & 45.49 & 82.89 & 79.79 \\
CODA-Prompt & 91.50 & 86.16 & 81.63 & 76.98 & 78.48 & 73.40 & 75.00 & 70.02 & 62.90 & 52.27 & 86.16 & 86.34 \\
SimpleCIL & 82.31 & 76.21 & 65.84 & 61.28 & 67.06 & 61.28 & 67.58 & 61.28 & 59.67 & 49.44 & 85.42 & 83.61 \\
APER + Adapter & 90.94 & 85.75 & 79.43 & 74.02 & 79.17 & 72.70 & 77.66 & 70.98 & 59.89 & 49.51 & 85.42 & 83.59 \\
APER + SSF & 89.81 & 83.92 & 79.82 & 74.13 & 78.93 & 72.25 & 75.60 & 68.53 & 63.33 & 52.60 & 86.75 & 84.51 \\
InfLoRA & 90.51 & 85.05 & 81.81 & 76.95 & 81.39 & 75.32 & 78.87 & 72.60 & 60.92 & 49.20 & 88.43 & 88.54 \\
SD-LoRA & 92.05 & 87.26 & 83.01 & 79.15 & 82.59 & 78.17 & 79.80 & 75.13 & 64.95 & 55.96 & 90.00 & 87.61\\
SEMA & 91.63 & 86.85 & 84.25 & 78.85 & 83.56 & 78.00 & 81.75 & 74.53 & 67.03 & 58.06 & 90.93 & 88.12 \\
\midrule
\textbf{StaR-MoE} & \textbf{92.69} & \textbf{88.78} & \textbf{86.09} & \textbf{81.62} & \textbf{84.96} & \textbf{79.43} & \textbf{82.90} & \textbf{76.28} & \textbf{71.60} & \textbf{61.75} & \textbf{94.43} & \textbf{93.79} \\
\bottomrule
\end{tabular}%
}
\end{table}

\subsection{Ablation Studies}
In this section, we conduct component-wise ablation studies on ImageNet-R (10 tasks) and ImageNet-A to evaluate SARA and ACR. We compare StaR-MoE with three variants: \textbf{Baseline}, which trains the newly added expert and router using only the classification loss; \textbf{w/o ACR}, which keeps only SARA; and \textbf{w/o SARA}, which keeps only ACR. The results are reported in Table~\ref{tab:ablation}.

As shown in Table~\ref{tab:ablation}, removing either SARA or ACR degrades the performance of StaR-MoE. The w/o ACR variant still substantially outperforms the baseline, confirming that SARA is crucial for preserving historical routing behavior and mitigating routing drift. The w/o SARA variant also improves over the baseline, indicating that ACR helps exploit the expanded expert capacity. However, ACR alone is consistently weaker than SARA alone, showing that capacity regularization cannot replace explicit routing alignment. The full StaR-MoE achieves the best results across all metrics, demonstrating the complementarity of SARA and ACR in expandable MoE.

\begin{table}[t]
\caption{Ablation studies on ImageNet-R (10 tasks) and ImageNet-A. We report average accuracy $\bar{\mathcal{A}}$ and last accuracy $\mathcal{A}_T$.}
\label{tab:ablation}
\centering
\footnotesize
\begin{tabular}{lcccccc}
\toprule
\multirow{2}{*}{Method} & \multicolumn{2}{c}{Components} & \multicolumn{2}{c}{ImageNet-R} & \multicolumn{2}{c}{ImageNet-A} \\
 & SARA & ACR & $\bar{\mathcal{A}}$ & $\mathcal{A}_T$ & $\bar{\mathcal{A}}$ & $\mathcal{A}_T$ \\
\midrule
Baseline & $\times$ & $\times$ & 80.82 & 75.20 & 66.56 & 57.87 \\
w/o ACR & \checkmark & $\times$ & 84.56 & 78.62 & 71.09 & 59.71\\
w/o SARA & $\times$ & \checkmark & 83.00 & 76.18 & 68.76 & 59.12 \\
\midrule
\textbf{StaR-MoE} & \checkmark & \checkmark & \textbf{84.96} & \textbf{79.43} & \textbf{71.60} & \textbf{61.75}\\
\bottomrule
\end{tabular}
\end{table}

\begin{figure}[t]
    \centering
    \begin{subfigure}[t]{0.32\linewidth}
        \centering
        \includegraphics[width=\linewidth,height=3.4cm,keepaspectratio]{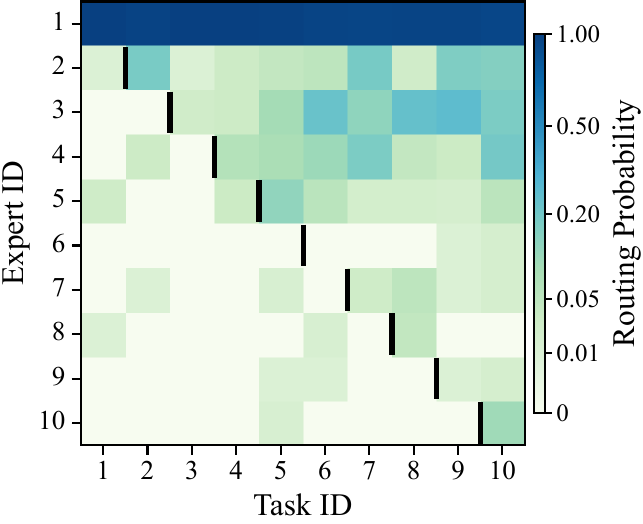}
        \caption{Routing stability visualization.}
        \label{fig:routing_stability}
    \end{subfigure}
    \hfill
    \begin{subfigure}[t]{0.32\linewidth}
        \centering
        \includegraphics[width=\linewidth,height=3.4cm,keepaspectratio]{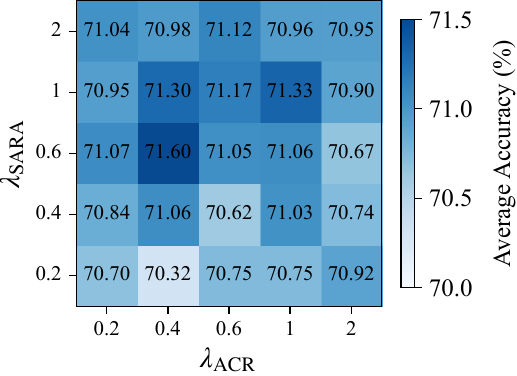}
        \caption{Parameter robustness.}
        \label{fig:hyperparameter_sensitivity}
    \end{subfigure}
    \hfill
    \begin{subfigure}[t]{0.32\linewidth}
        \centering
        \includegraphics[width=\linewidth,height=3.4cm,keepaspectratio]{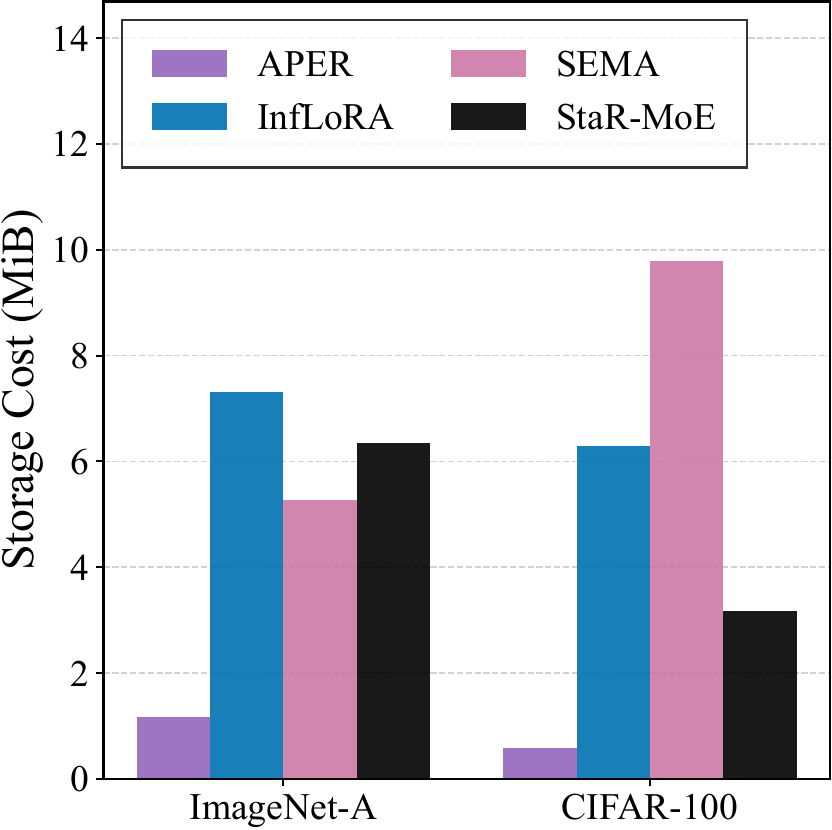}
        \caption{Storage cost.}
        \label{fig:memory_cost}
    \end{subfigure}
    \caption{
    Further analysis.
    (a) Task-wise routing probabilities in StaR-MoE on ImageNet-A. Low probabilities in the lower-left region indicate that later experts are rarely activated for earlier tasks.
    (b) Robustness analysis of $\lambda_{\mathrm{SARA}}$ and $\lambda_{\mathrm{ACR}}$ on ImageNet-A, showing robustness to the strengths of routing alignment and capacity regularization.
    (c) Extra storage cost of different methods.
    }
    \label{fig:further_analysis}
    \vspace{-0.5em}
\end{figure}

\subsection{Further Analysis}
\textbf{Routing stability visualization.}
We visualize the task-wise routing probabilities of StaR-MoE on ImageNet-A, consistent with the dataset used in the routing drift illustration in Figure~\ref{fig:router_drift}. To better reveal small routing probabilities, we apply a non-uniform color scale to the heatmap. As shown in Figure~\ref{fig:routing_stability}, the lower-left region has consistently low probabilities, indicating that later experts are rarely activated for earlier tasks. In contrast to the routing drift observed in Figure~\ref{fig:router_drift}, StaR-MoE suppresses the activation of newly added experts on old tasks. This visualization validates that StaR-MoE mitigates routing drift and preserves historical routing behavior.

\textbf{Parameter robustness.}
We analyze the sensitivity of StaR-MoE to $\lambda_{\mathrm{SARA}}$ and $\lambda_{\mathrm{ACR}}$, which control the strengths of routing alignment and capacity regularization, respectively. Specifically, we vary both coefficients within $\{0.2, 0.4, 0.6, 1.0, 2.0\}$. As shown in Figure~\ref{fig:hyperparameter_sensitivity}, StaR-MoE maintains stable performance across a wide range of values, indicating that the method is robust to these coefficients.

\textbf{Storage cost.}
We further compare the extra storage cost of StaR-MoE with representative CIL methods. As shown in Figure~\ref{fig:memory_cost}, StaR-MoE incurs moderate storage overhead on ImageNet-A and requires substantially less storage than SEMA and InfLoRA on CIFAR-100. These results show that StaR-MoE preserves routing stability without introducing excessive storage overhead.

\section{Conclusion}
\label{sec:conclusion}

In this paper, we have introduced Stable Routing for MoE (StaR-MoE), a routing-level framework for expandable MoE in CIL, motivated by the observation that continual expert expansion can induce routing drift. Specifically, newly added experts may alter old-class routing behavior and contribute to forgetting. To mitigate routing drift, StaR-MoE incorporates sensitivity-aware routing alignment to preserve old-class routing behavior by imposing stronger constraints on prediction-sensitive router layers. It further introduces asymmetric capacity regularization to encourage effective utilization of the expanded expert pool. Theoretical analysis connects SARA to an upper bound on aggregate routing drift, and extensive experiments validate StaR-MoE across multiple CIL benchmarks. In the future, extending StaR-MoE to multimodal continual learning settings remains a challenging direction, where stable routing must be maintained across diverse multimodal data streams.

\bibliographystyle{plainnat}
\bibliography{neurips_2026}

\newpage
\appendix

\section{Proof of Proposition \ref{prop:sara_drift_bound}}
\label{app:proof_propo_1}

We first bound the routing drift for a single MoE layer $\ell$. Following the notation in Sec.~\ref{sec:sara}, let $\hat{\boldsymbol{P}}_{c}^{\ell}$ denote the zero-padded target distribution, and let $\boldsymbol{P}^{\ell}(\tilde{\boldsymbol{z}}_{c}^{\ell})$ denote the current dense routing distribution on the synthesized router input $\tilde{\boldsymbol{z}}_{c}^{\ell}$. For the set of old classes $\mathcal{C}_{\mathrm{old}}$ with $N = |\mathcal{C}_{\mathrm{old}}|$, the layer-wise routing drift is defined as:
\begin{equation}
    \Delta^{\ell}
    =
    \sum_{c \in \mathcal{C}_{\mathrm{old}}}
    \mathbb{E}_{\tilde{\boldsymbol{z}}_{c}^{\ell}}
    \left[
    \left\|
    \hat{\boldsymbol{P}}_{c}^{\ell}
    -
    \boldsymbol{P}^{\ell}(\tilde{\boldsymbol{z}}_{c}^{\ell})
    \right\|_1
    \right].
\end{equation}
Since both distributions are defined over the same expanded expert set, their $L_1$ distance is twice the total variation (TV) distance:
\begin{equation}
    \left\|
    \hat{\boldsymbol{P}}_{c}^{\ell}
    -
    \boldsymbol{P}^{\ell}(\tilde{\boldsymbol{z}}_{c}^{\ell})
    \right\|_1
    =
    2\delta_{\mathrm{TV}}
    \left(
    \hat{\boldsymbol{P}}_{c}^{\ell},
    \boldsymbol{P}^{\ell}(\tilde{\boldsymbol{z}}_{c}^{\ell})
    \right).
\end{equation}
By Pinsker's inequality, for any two distributions $P$ and $Q$, $\delta_{\mathrm{TV}}(P,Q) \leq \sqrt{\frac{1}{2}D_{\mathrm{KL}}(P\parallel Q)}$. Applying this inequality to the zero-padded target distribution and the current dense routing distribution, and using Jensen's inequality $\mathbb{E}[\sqrt{X}] \leq \sqrt{\mathbb{E}[X]}$, we obtain:
\begin{equation}
\begin{aligned}
    \Delta^{\ell}
    &=
    \sum_{c \in \mathcal{C}_{\mathrm{old}}}
    2\mathbb{E}_{\tilde{\boldsymbol{z}}_{c}^{\ell}}
    \left[
    \delta_{\mathrm{TV}}
    \left(
    \hat{\boldsymbol{P}}_{c}^{\ell},
    \boldsymbol{P}^{\ell}(\tilde{\boldsymbol{z}}_{c}^{\ell})
    \right)
    \right] \\
    &\leq
    \sum_{c \in \mathcal{C}_{\mathrm{old}}}
    2\mathbb{E}_{\tilde{\boldsymbol{z}}_{c}^{\ell}}
    \left[
    \sqrt{
    \frac{1}{2}
    D_{\mathrm{KL}}
    \left(
    \hat{\boldsymbol{P}}_{c}^{\ell}
    \parallel
    \boldsymbol{P}^{\ell}(\tilde{\boldsymbol{z}}_{c}^{\ell})
    \right)}
    \right] \\
    &\leq
    \sqrt{2}
    \sum_{c \in \mathcal{C}_{\mathrm{old}}}
    \sqrt{
    \mathbb{E}_{\tilde{\boldsymbol{z}}_{c}^{\ell}}
    \left[
    D_{\mathrm{KL}}
    \left(
    \hat{\boldsymbol{P}}_{c}^{\ell}
    \parallel
    \boldsymbol{P}^{\ell}(\tilde{\boldsymbol{z}}_{c}^{\ell})
    \right)
    \right]
    }.
\end{aligned}
\end{equation}
To bound the summation over old classes, we use the Cauchy--Schwarz inequality. For non-negative values $x_c$, $\sum_{c=1}^{N}\sqrt{x_c} \leq \sqrt{N\sum_{c=1}^{N}x_c}$. Letting \(x_c\) be given by:
\begin{equation}
    x_c
    =
    \mathbb{E}_{\tilde{\boldsymbol{z}}_{c}^{\ell}}
    \left[
    D_{\mathrm{KL}}
    \left(
    \hat{\boldsymbol{P}}_{c}^{\ell}
    \parallel
    \boldsymbol{P}^{\ell}(\tilde{\boldsymbol{z}}_{c}^{\ell})
    \right)
    \right],
\end{equation}
we have:
\begin{equation}
    \sum_{c \in \mathcal{C}_{\mathrm{old}}}
    \sqrt{x_c}
    \leq
    \sqrt{N}
    \sqrt{
    \sum_{c \in \mathcal{C}_{\mathrm{old}}}
    \mathbb{E}_{\tilde{\boldsymbol{z}}_{c}^{\ell}}
    \left[
    D_{\mathrm{KL}}
    \left(
    \hat{\boldsymbol{P}}_{c}^{\ell}
    \parallel
    \boldsymbol{P}^{\ell}(\tilde{\boldsymbol{z}}_{c}^{\ell})
    \right)
    \right]
    }.
\end{equation}
The term inside the square root is exactly the single-layer routing alignment loss $\mathcal{L}_{\mathrm{align}}^{\ell}$. Therefore, we obtain:
\begin{equation}
    \Delta^{\ell}
    \leq
    \sqrt{2N\mathcal{L}_{\mathrm{align}}^{\ell}}.
\end{equation}

We now extend the result to the sensitivity-weighted routing drift across all MoE layers. By definition, we have:
\begin{equation}
    \Delta_{\mathrm{total}}
    =
    \sum_{\ell\in\mathcal{L}_{\mathrm{moe}}}
    \tilde{\alpha}^{\ell}\Delta^{\ell}.
\end{equation}
Substituting the layer-wise bound gives:
\begin{equation}
    \Delta_{\mathrm{total}}
    \leq
    \sqrt{2N}
    \sum_{\ell\in\mathcal{L}_{\mathrm{moe}}}
    \tilde{\alpha}^{\ell}
    \sqrt{\mathcal{L}_{\mathrm{align}}^{\ell}}.
\end{equation}
The interpolated sensitivity weights $\tilde{\alpha}^{\ell}$ are non-negative and satisfy $\sum_{\ell\in\mathcal{L}_{\mathrm{moe}}}\tilde{\alpha}^{\ell}=1$. Applying Jensen's inequality to the concave square-root function yields:
\begin{equation}
    \sum_{\ell\in\mathcal{L}_{\mathrm{moe}}}\tilde{\alpha}^{\ell}\sqrt{\mathcal{L}_{\mathrm{align}}^{\ell}}\leq\sqrt{\sum_{\ell\in\mathcal{L}_{\mathrm{moe}}}\tilde{\alpha}^{\ell}\mathcal{L}_{\mathrm{align}}^{\ell}}=\sqrt{\mathcal{L}_{\mathrm{SARA}}}.
\end{equation}
Thus, we obtain:
\begin{equation}
    \Delta_{\mathrm{total}}\leq\sqrt{2N\mathcal{L}_{\mathrm{SARA}}}.
\end{equation}
If the SARA objective satisfies $\mathcal{L}_{\mathrm{SARA}}\leq\xi$, then we have:
\begin{equation}
    \Delta_{\mathrm{total}}
    \leq
    \sqrt{2N\xi}.
\end{equation}
This proves Proposition~\ref{prop:sara_drift_bound}.

\section{Smooth Load Estimator}
\label{app:smooth_load}

In ACR, directly using the hard Top-$k$ assignment indicator $\mathbb{I}[j\in\mathcal{K}]$ to compute expert load is non-differentiable. Therefore, we use a smooth estimator $\mathcal{S}(j\in\mathcal{K}\mid\boldsymbol{z})$ to approximate the probability that expert $j$ is selected into the Top-$k$ expert set~\citep{shazeer2017outrageously}.

Given the router input $\boldsymbol{z}$, the routing logit of expert $j$ is:
\begin{equation}
    g_j(\boldsymbol{z})=\left(\boldsymbol{z}\mathbf{W}_{\mathrm{router}}\right)_j .
\end{equation}
Let $\tau_k(\boldsymbol{z}, j)$ denote the $k$-th highest logit value among all other experts excluding $j$ for router input $\boldsymbol{z}$, and let $\sigma_j$ denote the smoothing scale. The smooth selection probability is computed as:
\begin{equation}
    \mathcal{S}\left(j\in\mathcal{K}\mid\boldsymbol{z}\right)=\Phi\left(\frac{g_j(\boldsymbol{z})-\tau_k(\boldsymbol{z},j)}{\sigma_j}\right),
\end{equation}
where $\Phi(\cdot)$ is the cumulative distribution function of the standard normal distribution. This differentiable estimator allows gradients from the load regularization term to be propagated to the router.

Using this estimator, the batch-level load of expert $j$ is computed as:
\begin{equation}
    L_j=\sum_{\boldsymbol{z}\in\mathcal{B}}\mathcal{S}\left(j\in \mathcal{K}\mid\boldsymbol{z}\right),
\end{equation}
which measures the total selection strength of expert $j$ over the current-task router inputs.

\section{StaR-MoE Algorithm}
\label{app:alg}
We present the detailed procedure in Algorithm~\ref{alg:star_moe}.
\begin{algorithm}[t]
   \caption{StaR-MoE for Continual Learning}
   \label{alg:star_moe}
\begin{algorithmic}[1]
   \State \textbf{Input:} Sequential Tasks $\{1,\dots,T\}$, Pre-trained ViT Backbone, Hyperparameters $\lambda_{\mathrm{SARA}},\lambda_{\mathrm{ACR}},\gamma$.
   \State \textbf{Initialize:} Frozen ViT Backbone, Initial Experts $\mathcal{E}$, Routers $\mathcal{R}$, and Empty Statistics $\mathcal{M}\leftarrow\emptyset$.
   \For{Task $t=1$ \textbf{to} $T$}
      \If{$t>1$}
         \State Expand $\mathcal{E}$ with new experts and $\mathcal{R}$ with new router parameters.
         \State Compute layer-wise sensitivity weights $\tilde{\alpha}$ via a single mini-batch from $\mathcal{D}_{t}$.
      \EndIf
      \For{each training epoch and mini-batch $(x,y)$ in $\mathcal{D}_t$}
         \State $\mathcal{L}_{\mathrm{cur}}\leftarrow$ Compute Cross-Entropy Loss via Eq.~(\ref{eq:cur_loss})
         \If{$t>1$}
             \State $\mathcal{L}_{\mathrm{SARA}}\leftarrow$ Routing Alignment using $\tilde{\boldsymbol{\alpha}}$ via Eq.~(\ref{eq:sara_loss})
             \State $\mathcal{L}_{\mathrm{ACR}}\leftarrow$ Encourage Capacity Utilization via Eq.~(\ref{eq:acr_loss})
         \Else
             \State $\mathcal{L}_{\mathrm{SARA}}\leftarrow0$, $\mathcal{L}_{\mathrm{ACR}}\leftarrow0$
         \EndIf
         \State Update by minimizing $\mathcal{L}_{\mathrm{total}}=\mathcal{L}_{\mathrm{cur}}+\lambda_{\mathrm{SARA}}\mathcal{L}_{\mathrm{SARA}}+\lambda_{\mathrm{ACR}}\mathcal{L}_{\mathrm{ACR}}$
      \EndFor
      \If{$t<T$}
         \State Update statistics $\mathcal{M}$ using $\mathcal{D}_t$.
      \EndIf
   \EndFor
\end{algorithmic}
\end{algorithm}

\section{Experimental Setups and Implementation Details} 
\label{app:detail}
Following established protocols~\citep{smith2023coda, wu2025sdlora}, we adopt ViT-B/16~\citep{dosovitskiy2021image} pre-trained on ImageNet-21K and fine-tuned on ImageNet-1K as our backbone. For fair comparison, the backbone remains frozen across all methods. In addition to the supervised backbone, we evaluate a self-supervised ViT-B/16 pre-trained with DINO~\citep{caron2021emerging}. Following SEMA~\citep{wang2025self}, we apply expert expansion only to the last six transformer layers, while each of the first six layers uses a single adapter without expansion. In the MoE layers, experts learned from previous tasks are frozen during subsequent training, while newly added experts and all router parameters are updated.

For consistent and fair comparison, we use the same class order and data partitions for all methods. Following \citet{rebuffi2017icarl, zhou2024class}, we randomly shuffle the class order using the seed $1993$ before partitioning classes into incremental tasks. The resulting task sequence and the corresponding training and testing splits are kept identical across all compared methods.

For baselines such as SimpleCIL, APER, InfLoRA, SD-LoRA, and SEMA, we utilize their official implementations. For prompting-based methods, namely L2P, DualPrompt, and CODA-Prompt, we adopt the open-source implementations from the PILOT toolbox~\citep{sun2023pilot}. All baseline methods follow the hyperparameter configurations recommended in their original publications or official implementations.

StaR-MoE is optimized using Adam~\citep{kingma2014adam} with $\beta_1 = 0.9$ and $\beta_2 = 0.999$, where the learning rate decays with cosine annealing. The training epochs are set to 30 for ImageNet-R, 10 for CIFAR-100, 15 for ImageNet-A, and 30 for VTAB. The adapter projection dimension $r$ is set to 8 for ImageNet-A and the 20-task ImageNet-R setting, and 16 for all other benchmarks. We consistently set the number of active experts $k=2$ across all datasets. For the method-specific hyperparameters of StaR-MoE, we split the original training set into training and validation subsets with a ratio of $4{:}1$, and select hyperparameters according to the validation performance without using the official test results. We set $\lambda_{\mathrm{SARA}}=0.6$, $\lambda_{\mathrm{ACR}}=0.4$, and $\gamma=0.5$ as the default configuration across datasets. Once the hyperparameters are fixed, the official test set is used exclusively for final evaluation. All experiments are conducted on NVIDIA GeForce RTX 4090 GPUs with 24GB of memory, using PyTorch 2.0.1.

\begin{figure}[t]
  \centering
  \includegraphics[width=0.9\textwidth]{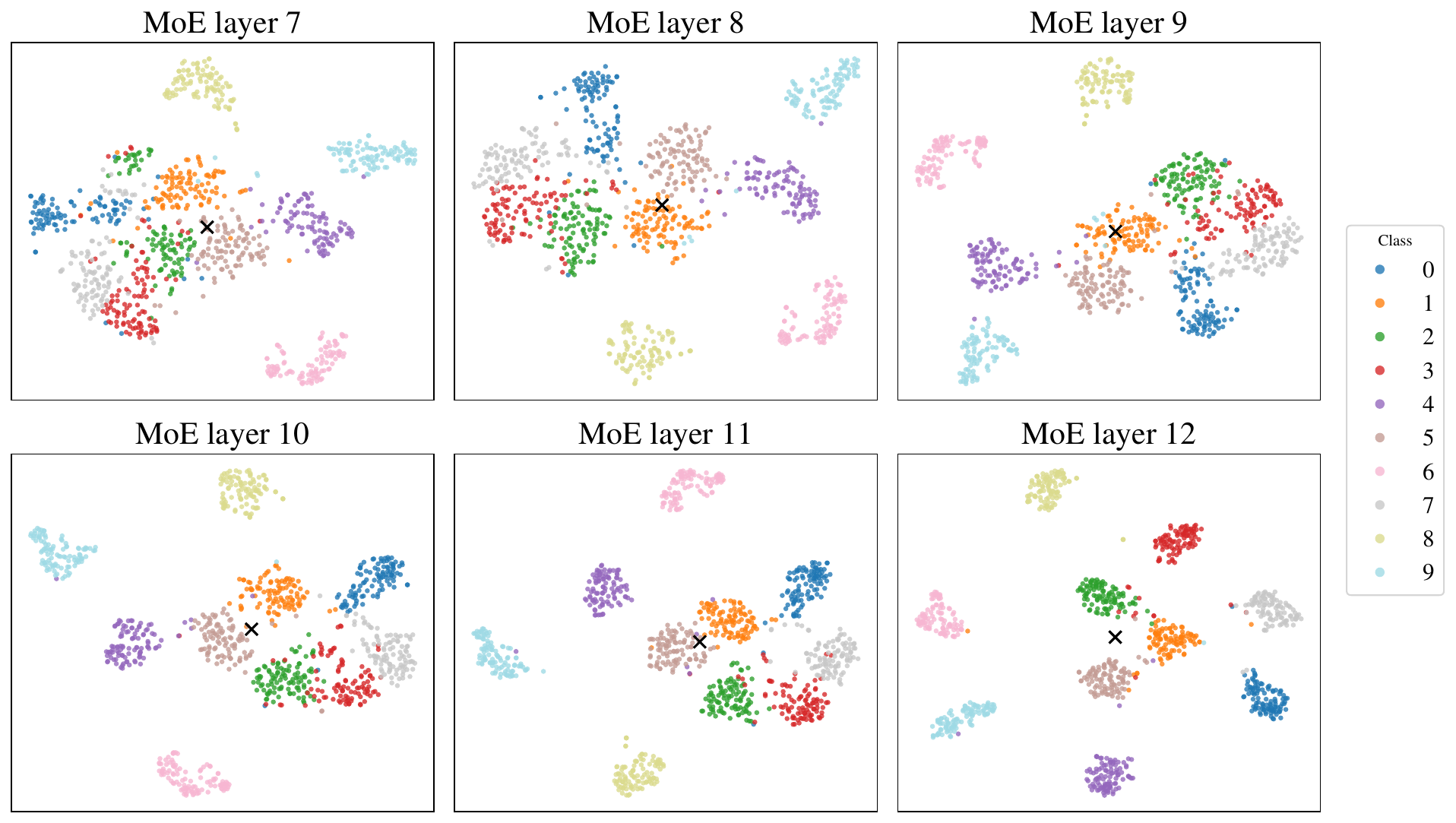} 
    \caption{
    t-SNE visualization of router-input distributions across MoE layers on VTAB.
    Samples from different classes exhibit compact local structures in the router-input space, providing empirical motivation for class-wise statistics.
    }
  \label{fig:router_input_visualization}
\end{figure}

\section{Additional Experiment Results}
\label{app:add_experiment}
\subsection{Visualization of Router-Input Distributions}
\label{app:router_input_visualization}

We visualize the router-input distributions across different MoE layers using t-SNE~\citep{van2008visualizing}, as shown in Figure~\ref{fig:router_input_visualization}. Router inputs from the same class generally exhibit compact local structures, suggesting that router inputs retain class-related information across layers. This observation provides empirical motivation for using class-wise statistics to approximate old-class router-input distributions in SARA.

\subsection{Effect of Sensitivity-Aware Weighting}
\label{app:sensitivity_weighting}
We further analyze the effect of the sensitivity-aware weighting strategy in SARA on ImageNet-A. All variants keep the same training setting, and differ only in the layer-wise alignment weights. We compare the proposed sensitivity-aware weighting with uniform alignment, which assigns equal weights to all MoE layers. As shown in Table~\ref{tab:sensitivity_weighting}, sensitivity-aware weighting outperforms uniform alignment, suggesting that assigning larger alignment weights to more sensitive layers provides a more effective constraint on routing drift.

\begin{table}[t]
\centering
\caption{
Effect of sensitivity-aware weighting in SARA on ImageNet-A.
Uniform assigns equal weights to all MoE layers.
}
\label{tab:sensitivity_weighting}
\begin{tabular}{lcc}
\toprule
Weighting Strategy & $\bar{\mathcal{A}}$ & $\mathcal{A}_T$ \\
\midrule
Uniform & $71.22$ & $61.16$ \\
Sensitivity-aware & \textbf{71.60} & \textbf{61.75} \\
\bottomrule
\end{tabular}
\end{table}

\begin{figure}[t]
    \centering
    \begin{subfigure}[t]{0.48\linewidth}
        \centering
        \includegraphics[width=0.98\linewidth]{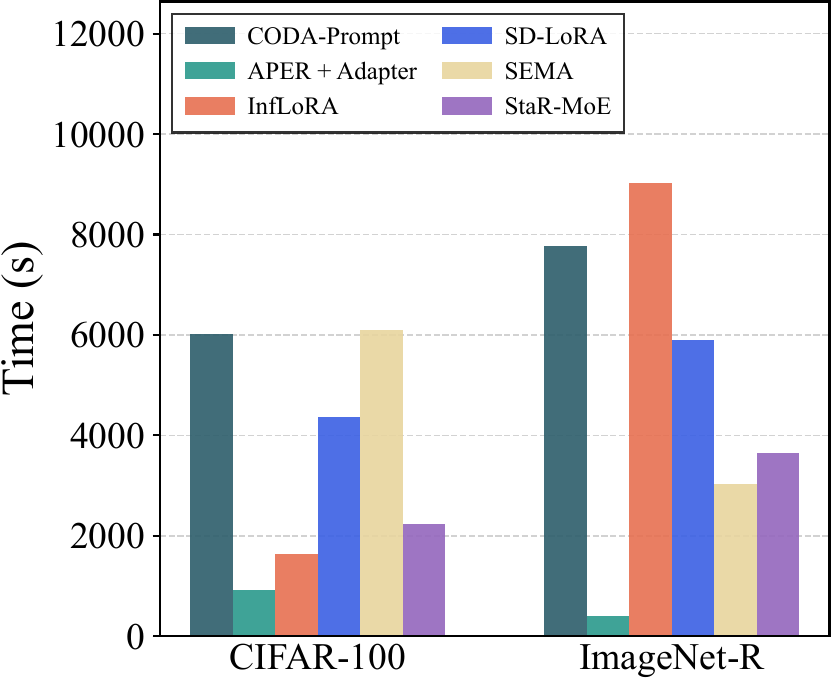}
        \caption{Running time comparison.}
        \label{fig:run_time}
    \end{subfigure}
    \hfill
    \begin{subfigure}[t]{0.48\linewidth}
        \centering
        \includegraphics[width=0.85\linewidth]{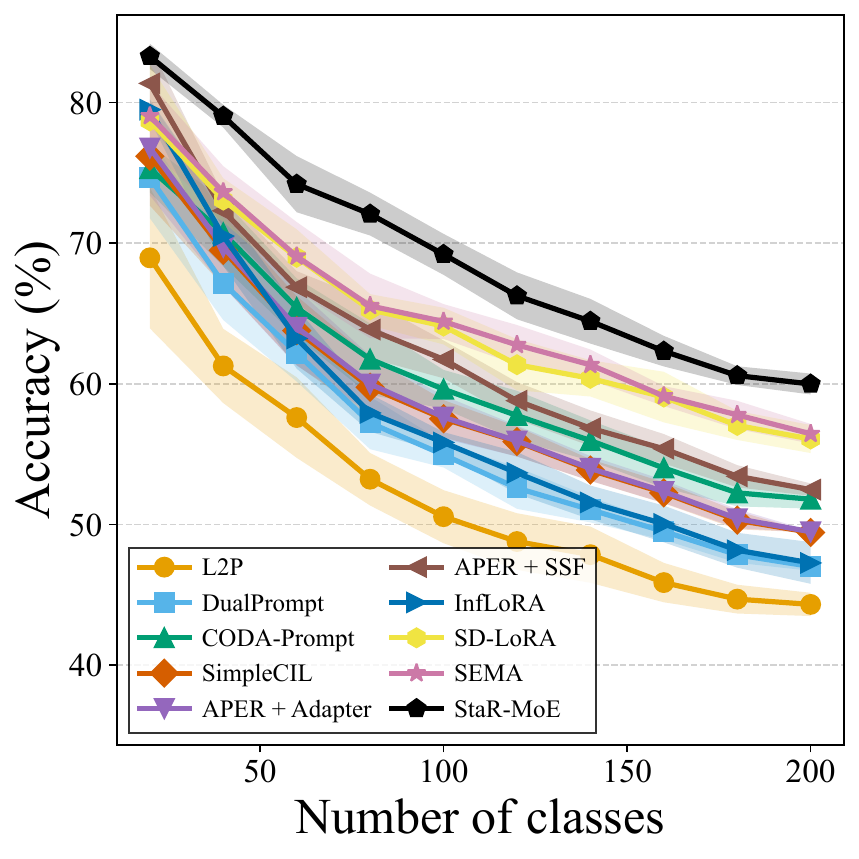}
        \caption{Multi-seed evaluation.}
        \label{fig:multi_runs}
    \end{subfigure}
    \caption{
    Additional experimental results.
    (a) Running time comparison on CIFAR-100 and ImageNet-R.
    \textbf{StaR-MoE maintains competitive training efficiency while achieving the best overall performance.}
    (b) Experimental results with different random seeds on ImageNet-A.
    StaR-MoE consistently outperforms the compared methods by a clear margin, demonstrating strong statistical stability.
    }
    \label{fig:additional_results}
\end{figure}

\subsection{Running Time Comparison}
\label{app:running_time}
To further evaluate the practical efficiency of StaR-MoE, we compare the training time of different methods on CIFAR-100 and ImageNet-R (10 tasks). All experiments are conducted under the same hardware environment using NVIDIA GeForce RTX 4090 GPUs. As shown in Figure~\ref{fig:run_time}, StaR-MoE maintains competitive training efficiency on both benchmarks while achieving the best overall performance in the main experiments, indicating that the proposed routing stabilization strategy does not introduce excessive computational overhead.

\subsection{Experimental Results with Different Seeds}
Following~\citet{rebuffi2017icarl}, the experiments in the main text use the random seed $1993$ to randomize the class order. To further evaluate the statistical stability of StaR-MoE, we repeat the experiments on ImageNet-A with multiple random seeds, namely $\{1993, 1994, 1995, 1996, 1997\}$. These repeated trials yield five sets of incremental results for each method, from which we compute the mean and standard deviation reported in Figure~\ref{fig:multi_runs}. As shown in the figure, StaR-MoE consistently achieves superior performance across different random seeds, indicating strong statistical stability.

\subsection{Impact of MoE Design Choices}
\label{app:moe_design_analysis}

\begin{figure}[t]
    \centering
    \begin{subfigure}[t]{0.48\linewidth}
        \centering
        \includegraphics[width=0.8\linewidth]{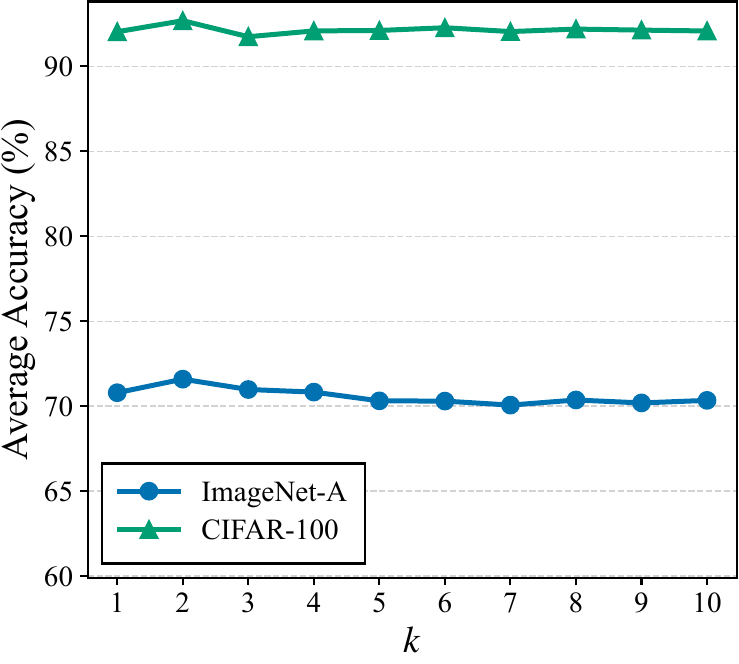}
        \caption{Number of active experts.}
        \label{fig:topk_sensitivity}
    \end{subfigure}
    \hfill
    \begin{subfigure}[t]{0.48\linewidth}
        \centering
        \includegraphics[width=0.8\linewidth]{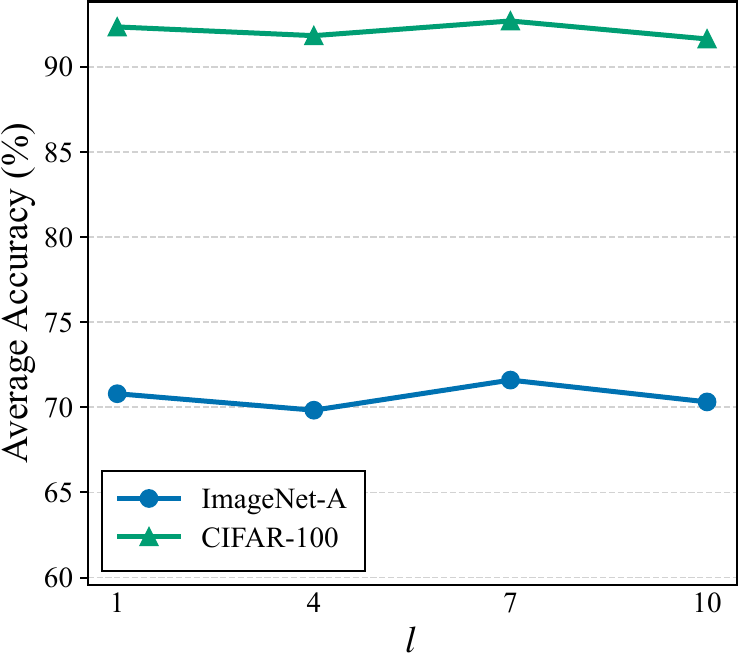}
        \caption{MoE layer placement.}
        \label{fig:moe_layer_placement}
    \end{subfigure}
    \caption{
    Impact of MoE design choices.
    (a) Effect of the number of active experts $k$.
    (b) Effect of the MoE starting layer $l$, where expert expansion is applied from layer $l$ to the last transformer layer.
    }
    \label{fig:moe_design_analysis}
\end{figure}

We further analyze two MoE design choices in StaR-MoE: the number of active experts $k$ and the placement of MoE layers. Figure~\ref{fig:moe_design_analysis} presents representative results on CIFAR-100 and ImageNet-A. For the number of active experts, StaR-MoE achieves relatively stable performance across different choices of $k$, and $k=2$ provides a favorable trade-off between routing specialization and capacity utilization. For MoE layer placement, we vary the starting layer $l$ of expert expansion, where expansion is applied from layer $l$ to the last transformer layer. The setting $l=7$ corresponds to expanding the last six transformer layers and achieves a practical trade-off between performance and expansion scope. We therefore use this setting as the default configuration.

\subsection{Varying the Pre-trained Model}
\label{app:dino}
To assess the generality of StaR-MoE, we perform experiments using a ViT-B/16 backbone pre-trained with the self-supervised method DINO~\citep{caron2021emerging} on ImageNet-R (10 tasks). As shown in Table~\ref{tab:dino_inr10}, StaR-MoE demonstrates improved performance over the baselines, indicating its effectiveness across different backbones.

\begin{table}[t]
\caption{Performance comparison on ImageNet-R (10 tasks) with DINO. We report average accuracy $\bar{\mathcal{A}}$ and last accuracy $\mathcal{A}_T$.}
\label{tab:dino_inr10}
\centering
\begin{tabular}{lcc}
    \toprule
    \multirow{2}{*}{Method} & \multicolumn{2}{c}{ImageNet-R} \\ 
        & $\bar{\mathcal{A}}$ & $\mathcal{A}_T$ \\
    \midrule
    L2P & $68.77$ & $61.94$ \\
    DualPrompt & $67.65$ & $60.40$ \\
    CODA-Prompt & $72.20$ & $64.63$ \\
    InfLoRA & $76.40$ & $67.91$ \\
    SD-LoRA & $64.61$ & $69.12$ \\
    SEMA & $72.55$ & $64.10$ \\
    \midrule
    \textbf{StaR-MoE} & \textbf{77.46} & \textbf{69.72} \\
    \bottomrule
\end{tabular}
\end{table}

\subsection{Compatibility with Selective Expert Expansion}
\label{app:selective_expansion}

We further evaluate StaR-MoE with a selective expansion policy inspired by SEMA~\citep{wang2025self}. In this variant, denoted as StaR-MoE-SE, new experts are added according to the selective expansion criterion in SEMA, while SARA and ACR remain unchanged. As shown in Table~\ref{tab:selective_expansion}, StaR-MoE-SE reduces the total number of expanded experts compared with the default task-wise expansion strategy, with only a moderate drop in accuracy. Importantly, the variant still outperforms SEMA, especially on VTAB, suggesting that routing stabilization remains beneficial even under selective expert growth.

\begin{table}[t]
\centering
\caption{Compatibility with selective expert expansion. Expert Reduction is computed relative to the default expansion of StaR-MoE.}
\label{tab:selective_expansion}
\begin{tabular}{llcccc}
\toprule
Dataset & Method & Expansion & Expert Reduction & $\bar{\mathcal{A}}$ & $\mathcal{A}_T$ \\
\midrule
\multirow{3}{*}{VTAB}
& SEMA & selective & -- & 90.93 & 88.12 \\
& StaR-MoE & task-wise & -- & 94.43 & 93.79\\
& StaR-MoE-SE & selective & 30.0\% & 93.38 & 92.25 \\
\midrule
\multirow{3}{*}{ImageNet-A}
& SEMA & selective & -- & 67.03 & 58.06 \\
& StaR-MoE & task-wise & -- & 71.60 & 61.75 \\
& StaR-MoE-SE & selective & 25.0\% & 69.33 & 58.92 \\
\bottomrule
\end{tabular}
\end{table}

\section{Storage and Parameter Overhead}
\label{app:storage_overhead}

\subsection{Storage Cost Discussion}
\label{app:storage_cost_discussion}
We report the method-specific extra storage cost in Figure~\ref{fig:memory_cost}. The frozen backbone and common components shared by all methods are excluded from this comparison. For APER, the extra storage comes from the prototypes maintained for its prototype-based classifier. For InfLoRA, it comes from the gradient-subspace bases maintained to keep LoRA updates orthogonal to old-task gradient subspaces. For SEMA, it corresponds to the additional expansion-related storage required by its dynamic model expansion. For StaR-MoE, it comes from the class-wise statistics used by SARA.

\subsection{Learnable Parameter Overhead}
We report the number of accumulated learnable parameters after the final task in Table~\ref{tab:param_overhead}. The frozen ViT backbone is excluded for all methods, since it is shared across compared methods. Among the compared baselines, DualPrompt and CODA-Prompt introduce task-dependent prompt parameters, while SD-LoRA introduces task-dependent LoRA components; these parameters grow linearly with the number of tasks. SEMA does not expand at every task, but its adapter modules can still increase when selective expansion is triggered. Therefore, we report the accumulated method-specific learnable parameters after completing all tasks. StaR-MoE introduces a moderate amount of learnable parameters among representative CIL methods. In particular, its parameter overhead remains much smaller than CODA-Prompt and SD-LoRA.

\begin{table}[t]
    \centering
    \caption{Learnable parameter comparison on ImageNet-R (10 tasks). Learnable Param. denotes the accumulated learnable parameters after the final task, excluding the frozen backbone.}
    \label{tab:param_overhead}
    \resizebox{0.45\linewidth}{!}{
        \begin{tabular}{lc}
            \toprule
            Method & Learnable Param. (M)\\
            \midrule
            L2P             & 0.05 \\
            DualPrompt      & 0.25 \\
            CODA-Prompt     & 3.84 \\
            APER + Adapter  & 1.48 \\
            InfLoRA         & 3.69 \\
            SD-LoRA         & 3.69 \\
            SEMA            & 1.35 \\
            \midrule
            StaR-MoE        & 1.88 \\
            \bottomrule
        \end{tabular}
    }
\end{table}



\end{document}